\pdfoutput=1
\documentclass{article} %
\usepackage{iclr2027_conference,times}

\usepackage{amsmath,amsfonts,bm}

\def\figref#1{figure~\ref{#1}}

\def\secref#1{section~\ref{#1}}

\def\eqref#1{equation~\ref{#1}}

\def\algref#1{algorithm~\ref{#1}}

\def\1{\bm{1}}

\DeclareMathAlphabet{\mathsfit}{\encodingdefault}{\sfdefault}{m}{sl}
\SetMathAlphabet{\mathsfit}{bold}{\encodingdefault}{\sfdefault}{bx}{n}

\def\eqref#1{Eq.~\ref{#1}}    
\def\figref#1{Fig.~\ref{#1}}  
\def\secref#1{Sec.~\ref{#1}}  
\def\algref#1{Alg.~\ref{#1}}  
\def\tabref#1{Tab.~\ref{#1}}  
\def\appref#1{App.~\ref{#1}}  

\usepackage{amsmath,amssymb}
\usepackage{booktabs}
\usepackage{array}
\usepackage{multirow}
\usepackage{graphicx}
\usepackage{xcolor}
\usepackage{colortbl}          %
\colorlet{oursrow}{white}
\usepackage{xspace}
\usepackage{algorithm}
\usepackage{algorithmic}
\usepackage{enumitem}
\usepackage{verbatim}   %
\usepackage{microtype}
\usepackage{hyperref}
\usepackage{url}
\definecolor{cvprblue}{rgb}{0.21,0.49,0.74}   %
\hypersetup{colorlinks=true, citecolor=cvprblue, linkcolor=red, urlcolor=black, breaklinks=true}

\newcommand{\method}{\textsc{Solo}\xspace}                 %
\newcommand{\setting}{fully test-time adaptation\xspace}   %
\newcommand{\Setting}{Fully test-time adaptation\xspace}   %
\newcommand{\topk}{top-$K$ self-distillation\xspace}
\newcommand{\hind}{hindsight relabeling\xspace}
\newcommand{\Hind}{Hindsight relabeling\xspace}            %
\newcommand{\uitars}{UI-TARS-7B\xspace}
\newcommand{\qwenvl}{Qwen3-VL-8B\xspace}

\newcommand{\wa}{WebArena\xspace}
\newcommand{\vwa}{VisualWebArena\xspace}
\newcommand{\mw}{MobileWorld\xspace}

\title{One Rollout Is All You Get: \\ Fully Test-Time Adaptation for GUI Agents}

\author{\parbox[t]{\dimexpr\textwidth-2\tabcolsep\relax}{\normalfont\raggedright
{\bfseries
Ziqiang~Wang$^{1,2}$ \quad Li~Gu$^{1,2}$ \quad Zhixiang~Chi$^{3}$ \quad
Linlian~Jiang$^{1,2}$ \quad Zihuan~Jiang$^{3}$ \quad Linqiang~Guo$^{1}$ \quad
Siobhan~Reid$^{1,2}$ \quad Zhi~Liu$^{4}$ \quad Yang~Wang$^{1,2}$}\\[4pt]
$^{1}$Concordia~University \quad $^{2}$Mila~--~Qu\'ebec~AI~Institute \quad
$^{3}$University~of~Toronto \quad $^{4}$Shanghai~University\\[2pt]
Correspondence: \texttt{ziqiang.wang@mail.concordia.ca}}}

\iclrfinalcopy %
\begin{document}

\maketitle
\lhead{Preprint}

\begin{abstract}
GUI agents are deployed with frozen weights and discard everything they experience on the job.
Existing ways to update an agent's weights assume something deployment withholds: ground truth, rollouts beyond the single attempt (retries, samples, practice runs), or a learning phase other than deployment.
Because GUI actions can be irreversible, a deployed agent gets one attempt per task occurrence, in arrival order, and every attempt counts.
No ground truth is available at any point.
We define \emph{\setting} for GUI agents by these constraints and pair it with a minimal weight-space method, \method.
Auxiliary models read each episode: a judge selects the episodes it deems successful, and a proposer--verifier pair relabels a failed episode's prefix with the subtask that prefix completed.
Admitted episodes enter a short sliding window, and each admission updates a small adapter by \topk on the agent's own predictions, provided the window holds a judged success.
On recurring task streams built from \wa, \vwa and \mw, \method improves on the frozen agent with both \uitars and \qwenvl, by three to six points of success rate, and exceeds two in-setting memory methods on the web streams.

\end{abstract}

\section{Introduction}
\label{sec:intro}

A GUI agent that books flights, files issues, or manages a calendar on a user's behalf \citep{qin2025uitars,scalecua2025,qwen3vl2025} ships with fixed weights, and from then on everything it experiences is discarded.
Yet the deployment stream is the distribution the agent is judged on, and for many deployments that distribution recurs.
A personal agent is the clearest case: a single user returns to the same apps and accounts, with the same errands coming back as new instances, the same expense form with a new amount, the same store with a new item.
An agent that cannot learn from this stream makes the same mistake on the tenth occurrence of a task as on the first.

Most methods for improving an agent after it ships rely on a resource that deployment does not provide, and \tabref{tab:paradigms} places them by the resources they use: ground-truth supervision, rollouts beyond the single attempt that counts, and a stage other than deployment at which to adapt.
Demonstration fine-tuning trains on labeled trajectories before deployment \citep{wu2024osatlas,qin2025uitars}.
Reinforcement learning and the self-evolution flywheels built on it learn from a reward, given by the environment or a trained reward model, with parallel rollouts, resets and an iterated training phase \citep{bai2024digirl,qi2025webrl,uigenie2025}.
Exploration memory is filled by practice runs before the attempts that count \citep{zhang2023appagent,sun2026magnet}, and retry-based improvement attempts a task again \citep{shinn2023reflexion,li2026jitrl,he2025evotest}.
Multi-rollout inference and adaptation draw many samples of each input to select among at inference or to train on at test time, or train on data derived from each test input \citep{snell2024scaling,yang2025gta1,zuo2025ttrl,akyurek2024ttt}.
A live deployment of a GUI agent provides none of these resources.

\begin{table}[t]
  \caption{\textbf{Where \setting sits among ways to improve a GUI agent.} Each row is placed by the resources its learning uses: ground-truth supervision (demonstrations, an environment check or a human; \emph{Varies}: some members use it, or a trained or VLM evaluator stands in), rollouts beyond the single attempt that counts, the stage at which adaptation happens, and the state retained across tasks. Our setting withholds the first two and admits only the deployment stage, with memory or weights as the retained state. The row marked $\dagger$ satisfies it. The last column lists representative works, and \secref{sec:related} discusses the rest.}
  \label{tab:paradigms}
  \centering
  \footnotesize
  \setlength{\tabcolsep}{3pt}
  \renewcommand{\arraystretch}{1.12}
  \begin{tabular}{>{\raggedright\arraybackslash}p{3.25cm}>{\centering\arraybackslash}p{1.45cm}>{\centering\arraybackslash}p{1.15cm}>{\centering\arraybackslash}p{2.0cm}>{\centering\arraybackslash}p{1.25cm}>{\raggedright\arraybackslash}p{3.6cm}}
    \toprule
    \bfseries Approach & \bfseries GT\newline supervision & \bfseries Extra\newline rollouts & \bfseries Adaptation\newline stage & \bfseries Retained\newline state & \bfseries Related work \\
    \midrule
    Demonstration SFT & Yes & No & \mbox{Pre-deployment} & Weights & \citet{wu2024osatlas,qin2025uitars} \\
    RL / self-evolution & Varies & Yes & \mbox{Pre-deployment} & Weights & \citet{bai2024digirl,qi2025webrl,uigenie2025} \\
    Exploration memory & Varies & Yes & \mbox{Pre-deployment} & Memory & \citet{zhang2023appagent,sun2026magnet} \\
    Retry-based improvement & Varies & Yes & Deployment & Memory & \citet{shinn2023reflexion,li2026jitrl,he2025evotest} \\
    Multi-rollout inference & No & Yes & Deployment & None & \citet{snell2024scaling,yang2025gta1} \\
    Multi-rollout adaptation & No & Yes & Deployment & Weights & \citet{akyurek2024ttt,zuo2025ttrl} \\
    Single-pass memory$^\dagger$ & No & No & Deployment & Memory & \citet{wang2024awm,mi2026darwinian,wang2025mobileagente} \\
    \midrule
    \rowcolor{oursrow} \textbf{\method (ours)} & \textbf{No} & \textbf{No} & \textbf{Deployment} & \textbf{Weights} & \\
    \bottomrule
  \end{tabular}
\end{table}

A GUI action can be irreversible: an email sent is sent, a purchase placed is placed.
An agent working in a real account cannot fork the world to sample many candidate action sequences, replay a task to try another route, or reset the environment between attempts.
It gets exactly one attempt per task occurrence, in the order the user issues tasks, and every attempt counts, because each is an errand the user wanted done.
We also assume no ground truth: no checker inspects the account, no annotator labels the outcome, and the user gives no feedback.
Whatever feedback exists must be computed from the episode itself by automated means, and unlike ground truth it can be wrong.

Following Tent \citep{wang2021tent}, which defined fully test-time adaptation for perception models by what it withholds (the correspondence is spelled out at the end of \secref{sec:setting}), we define \emph{\setting} for GUI agents by three constraints: no ground truth in any form, no rollouts beyond the single attempt per task occurrence (no retries, samples or exploration episodes), and no learning phase other than deployment, including practice runs before the tasks that count.
Any signal that automated models compute from the agent's own episodes is permitted, and so is any state that persists across the stream, external memory or weights alike.
What remains is an off-the-shelf agent, its tasks in arrival order, and one attempt per task occurrence.
Single-pass memory systems driven by a model judge \citep{wang2024awm,mi2026darwinian} already satisfy these constraints, and every other approach in \tabref{tab:paradigms} relies on at least one withheld resource.

The first design choice of our method, \method, is the learning signal.
Tent trusts the model's confidence and minimizes its entropy, but for a GUI agent entropy minimization leaves the success rate at the frozen level on \wa and collapses the agent on \vwa (\appref{app:signal}).
\method therefore lets auxiliary models read each episode.
A judge decides whether the episode completed its task \citep{pan2024autoeval} and thereby selects what to learn from.
For a judged failure, a proposer names a subtask that a prefix of the episode completed, as in hindsight relabeling \citep{andrychowicz2017her,zhang2023hir}, and an independent verifier checks the claim against the screenshots before the prefix is admitted.
None of these models generates an action or a token target: they decide which of the agent's own episodes to learn from and under which instruction.

\begin{figure}[t]
  \centering
  \includegraphics[width=\linewidth]{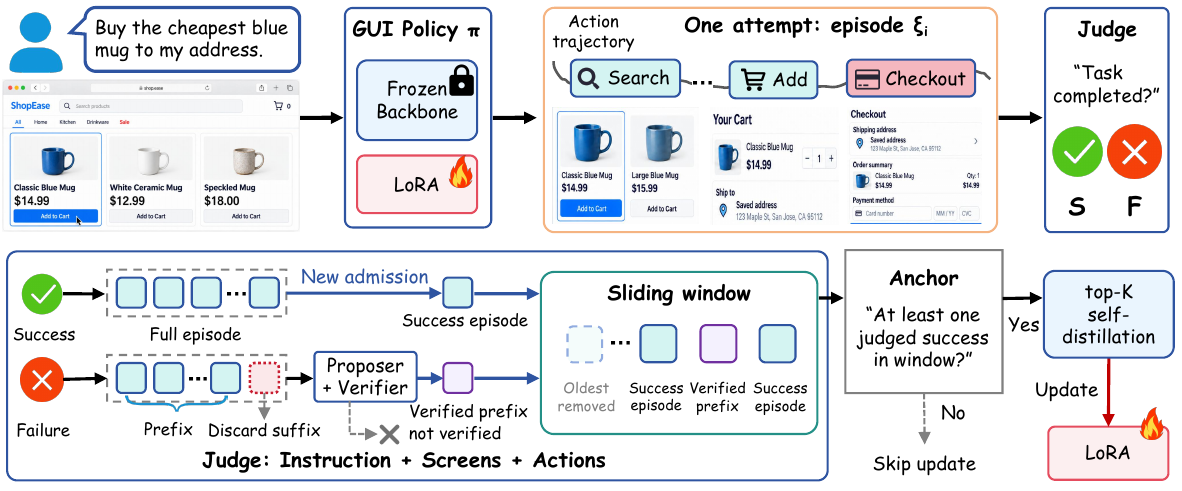}
\caption{\textbf{Fully test-time adaptation for GUI agents, and \method.}
Top: each task occurrence gets one attempt, and a judge decides from the instruction,
screens and actions whether it was completed. Bottom: a judged success enters
a sliding window as a full episode; for a judged failure, a proposer--verifier
pair relabels a completed prefix, which enters the same window. While the
window holds a judged success, one \topk step updates the LoRA adapter.}
  \label{fig:overview}
\end{figure}

A judged success enters a short sliding window of recent admissions, and an admitted relabeled prefix enters the same window in place of the failed episode.
Each admission triggers one small update of a low-rank adapter on the frozen backbone by \topk: the targets are the policy's own top-$K$ predictions at every position, so the executed tokens are never imitated outright.
Updates fire only while the window holds at least one judged success, so a relabeled prefix is trained only together with a success.
Apart from the window, the adapter is the only state carried across tasks (\figref{fig:overview}).

We make three contributions.
First, we formulate \setting for GUI agents by three constraints that deployment imposes (\secref{sec:setting}) and place the existing approaches to improving GUI agents against it (\tabref{tab:paradigms}).
Second, we propose \method, a minimal weight-space method for the setting, in which a judge and a proposer--verifier pair read each episode and the admitted episodes and relabeled prefixes train a small adapter by \topk (\secref{sec:method}).
Third, on recurring task streams we build in \wa, \vwa and \mw, with two open agents, \method improves on the frozen agent by three to six points of success rate and exceeds two in-setting memory methods on the web streams (\secref{sec:experiments}).

\section{Setting: fully test-time adaptation for GUI agents}
\label{sec:setting}

\par\noindent{\bf Deployment stream.}
A deployed agent faces tasks $\tau_1, \tau_2, \dots$ in an order it does not control.
A task $\tau=(g,e)$ pairs a natural-language instruction $g$ with the environment state $e$ in which it is issued.
We call each position in the stream an occurrence.
Tasks that share a template (the same errand with different arguments) recur along the stream, and some exact instances recur as well.
For each occurrence the agent $\pi_\theta$ produces one episode $\xi=(o_0,y_0,o_1,y_1,\dots,o_H)$ by observing the screen $o_t$, emitting a response $y_t$ that contains its reasoning and an action, and continuing until it declares the task done or exhausts the step budget $B(\tau)$.
Actions take effect in the environment, some of them irreversibly.
There is no separate evaluation set: the stream is the evaluation, and every episode counts.

\par\noindent{\bf What is withheld, what is permitted.}
\Setting for GUI agents is the problem of improving $\pi_\theta$ along this stream under three constraints.
\emph{No ground truth}: no labels, rewards, demonstrations, or human feedback are available for any task after deployment, and the evaluator that scores the stream is invisible to the learner.
\emph{One attempt per occurrence}: the agent produces exactly one episode for each $\tau_i$, in order, and that episode is the one that counts.
It may not retry, sample alternatives, branch or reset the environment, and it may not practice on tasks before they count.
\emph{Deployment only}: the backbone is an off-the-shelf checkpoint, no training phase precedes or interleaves the stream, and whatever learning happens draws only on episodes the agent has already produced.
Everything else is permitted.
In particular, any signal that automated models can compute from the agent's own episodes is admissible, such as a model judging whether an episode completed its task, and any state may persist across the stream, whether an external memory or the agent's own weights.

\par\noindent{\bf Protocol.}
Because the stream is the evaluation, performance is measured prequentially: each episode is scored as it is produced, and the score is the benchmark's own success criterion, which the learner never sees.
The quantity of interest is the difference in success rate between the adapting agent and the same agent frozen, both run on the identical stream in the identical order.
Success over the stream and its profile across rounds are the reported quantities.

\par\noindent{\bf In-setting neighbors.}
Stream-based memory systems already satisfy the three constraints (\tabref{tab:paradigms}, \secref{sec:related}), and we compare against two of them.
Workflow memory in its online form \citep{wang2024awm} processes test queries as a single stream and induces a workflow whenever a model-based evaluator judges the episode a success, and Darwinian Memory \citep{mi2026darwinian} runs the same tasks for several rounds while a self-verifier prunes and reinforces a memory of sub-task trajectories.
Both keep their experience in an external store and bring it back into the context or replay it.
The agent's weights never change.
\method occupies the other admissible choice of persistent state, the weights.

\par\noindent{\bf Relation to test-time adaptation of perception models.}
Tent defined fully test-time adaptation for perception models by withholding the labels and the training phase \citep{wang2021tent}: the model arrives already trained, may not revisit the data it was trained on or change how it was trained, and is left alone with its test stream.
Two of the three constraints above are its analogues: no ground truth, and no phase other than deployment.
The third has none: running a perception model on an input has no consequence and can be repeated after an update, whereas an occurrence here is acted upon once and cannot be replayed.
The one-attempt constraint encodes this difference.
In both settings a learning signal can only be computed from what the model has already produced, but what it produces differs, and so does the signal.
For a perception model it is the prediction it has just made, which is why minimizing its entropy, or training on its own confident predictions, are the canonical test-time signals.
For a GUI agent it is the episode it has just produced, and the counterpart signal is a reading of that episode: an automated model can estimate from the episode alone whether it completed its task, without ground truth and without a second attempt, so this signal is admissible here for the same reason confidence is admissible there.
Which signal to compute, and how to use it, is a method choice (\secref{sec:method}).

\section{Method: \method}
\label{sec:method}
\method adapts a small low-rank adapter along the stream (\figref{fig:overview}, \algref{alg:solo}).
Tasks are handled strictly in arrival order.
For each $\tau_i$ the agent produces one episode $\xi_i$ under its current adapter, and that episode is what the stream records.
Three auxiliary models then read $\xi_i$: a judge decides whether the task was completed, and for a judged failure a proposer and a verifier decide whether some prefix of the episode completed a subtask of its own.
A judged success is admitted as it stands, and a verified prefix is admitted under the subtask's instruction.
Admitted episodes enter a sliding window of the $W$ most recent admissions, and each admission triggers one gradient step of \topk on the window, applied to the adapter, provided the window holds at least one judged success.
Apart from the window, the adapter is the only state carried from one task to the next.

\par\noindent{\bf Signals and relabeling.}
None of the three auxiliary models generates an action or a token target.
We write $a_t$ for the action contained in the response $y_t$ and $\xi_{:k}=(o_0,y_0,\dots,o_k,y_k)$ for the prefix of an episode through step $k$.
The judge $J(g,\xi)\in\{0,1\}$ reads the instruction with the episode's screens and actions and decides whether the task was completed \citep{pan2024autoeval}.
For a judged failure, the proposer $G(g,\xi)$ reads the full episode and either abstains or names an instruction $g'$ that some prefix completed, together with the step $k$ at which it was completed, as in \hind \citep{andrychowicz2017her,zhang2023hir}.
The proposal is constrained: $g'$ must be a subtask, prerequisite or narrower version of $g$ about the same objects, and must be non-trivial and end on a resulting page or state.
Guards apply before verification: the prefix must hold at least two actions and must not be dominated by repeated actions, since a prefix that contains the agent's loops would teach the loops.
The verifier $V(g',a_{0:k},o_{k+1})$ then checks the claim independently, seeing only $g'$, the actions of the prefix and the screen that followed its last action, and accepts only if that screen shows $g'$ completed.
An accepted prefix $\xi_{:k}$ is admitted under $g'$ as an ordinary episode: it enters the same window with the same weight and is trained with the same objective over the same positions, reasoning included.
The branch is anchored: the window trains only while it holds at least one judged success, so a relabeled prefix is trained only in a window that also holds a judged success, and a prefix that leaves the window before a success arrives is never trained.
The judge thus decides what the agent learns from, and the proposer and the verifier decide under which instruction a failed prefix may count.

\par\noindent{\bf Objective: \topk.}
For an admitted episode $\xi$ with supervised positions $\mathcal{M}(\xi)$, let $\bar\theta$ denote the pre-update parameters with gradients stopped, $\mathcal{V}_K(t)$ the $K$ most probable tokens under $p_{\bar\theta}(\cdot\mid \xi_{<t})$, and
\begin{equation}
  \label{eq:topk}
  \mathcal{L}_K(\theta;\xi) \;=\; -\sum_{t\in\mathcal{M}(\xi)}\;\sum_{v\in\mathcal{V}_K(t)} q_t(v)\,\log p_\theta(v\mid \xi_{<t}),
  \qquad
  q_t(v) \;=\; \frac{p_{\bar\theta}(v\mid \xi_{<t})}{\sum_{u\in\mathcal{V}_K(t)} p_{\bar\theta}(u\mid \xi_{<t})}.
\end{equation}
The supervised positions are the tokens of the agent's own responses, reasoning and action alike, each conditioned on the observations and responses before it.
The targets $q_t$ come from the same forward pass with gradients stopped, so no second model and no cached teacher is involved.
First, the signals select episodes, not tokens.
A judged success is a verdict on the episode's outcome, and the objective never treats the executed token as correct: the target at every position is the truncated distribution of the pre-update policy itself.
Second, the gradient with respect to the logits at a position is $p_\theta-q_t$ on the support and $p_\theta$ off it, which at the pre-update point is proportional to the tail mass $1-\sum_{v\in\mathcal{V}_K(t)}p_{\bar\theta}(v)$.
Each step therefore moves probability from the tail onto the policy's own top-$K$ candidates, in proportion to their current probabilities and preserving their order.
Every response position is in the loss, but a position where the policy is already near-deterministic has almost no tail mass and contributes almost nothing, so the update concentrates where the policy is uncertain.
Third, $K$ interpolates between two extremes.
One-hot imitation of the executed tokens, the standard behavior-cloning loss, is not a self-distillation: its target puts all the probability on the executed token, so its gradient vanishes only when the policy has no alternative left at that position, and each update suppresses the alternatives the policy would need where the recorded action does not fit.
In our ablation the one-hot variant's episodes lengthen round by round, and by the third round three quarters of its \wa episodes run to the step budget (\secref{sec:anatomy}).
At the other extreme, $K=|\mathcal{V}|$ is a no-op.
A small $K$ keeps the update a self-distillation that leaves the policy's own ranking intact.

\par\noindent{\bf Update.}
All adaptation lives in a small adapter $\phi$ on the frozen backbone $\theta_0$, initialized with zero output (\appref{app:method}) so that the deployed policy $\pi_{\theta_0\oplus\phi}$ starts identical to the frozen agent and everything it becomes is attributable to the stream.
Each admission triggers one gradient step on the mean of \eqref{eq:topk} over the window, at a small step size.
Each episode takes part in at most $W$ updates and then leaves the window, and nothing is sampled from older episodes.
Per task, the cost is one judge call, at most two further calls for a judged failure, and at most one gradient step over $W$ episodes, and the adapter is about 0.2\% of the backbone.

\begin{algorithm}[t]\small
\caption{\method: \setting for a GUI agent}
\label{alg:solo}
\begin{algorithmic}[1]
\REQUIRE frozen backbone $\theta_0$; adapter $\phi\leftarrow 0$; judge $J$; proposer $G$; verifier $V$; window size $W$; support size $K$; step size $\eta$
\STATE $\mathcal{E}\leftarrow$ empty FIFO of capacity $W$
\FOR{each task $\tau_i=(g_i,e_i)$ in arrival order}
  \STATE $\xi_i \leftarrow$ \textsc{Rollout}$(\pi_{\theta_0\oplus\phi},\tau_i)$ \hfill $\triangleright$ one attempt; its outcome is the evaluation
  \STATE $b_i \leftarrow J(g_i,\xi_i)$ \hfill $\triangleright$ automated success judgment
  \IF{$b_i=1$}
    \STATE push $\xi_i$ onto $\mathcal{E}$
  \ELSE
    \STATE $(g',k)\leftarrow G(g_i,\xi_i)$ \hfill $\triangleright$ subtask completed by a prefix, or $\bot$
    \IF{$(g',k)\neq\bot$ \AND $V(g',\,a_{0:k},\,o_{k+1})$}
      \STATE push $(g',\xi_{i,:k})$ onto $\mathcal{E}$ \hfill $\triangleright$ relabeled prefix
    \ENDIF
  \ENDIF
  \IF{an episode was pushed \AND $\mathcal{E}$ holds a judged success}
    \STATE $\phi \leftarrow \phi - \eta\,\nabla_\phi \frac{1}{|\mathcal{E}|}\sum_{\xi\in\mathcal{E}} \mathcal{L}_K(\theta_0\oplus\phi;\xi)$ \hfill $\triangleright$ \eqref{eq:topk}, one gradient step; anchored
  \ENDIF
\ENDFOR
\end{algorithmic}
\end{algorithm}

\section{Experiments}
\label{sec:experiments}

\subsection{Setup and protocol}
\label{sec:setup}
\par\noindent{\bf Streams.}
We build three recurring streams, one per benchmark, and run every method on the identical stream in the identical order.
The \wa stream holds 108 task instances from 27 templates across four sites (an e-commerce store, its admin panel, a GitLab instance and a Reddit-style forum).
The \vwa stream holds 137 instances from 39 templates across a shopping site, a classifieds site and a forum.
The \mw stream holds 40 tasks spanning Gmail, Mastodon, device settings and native Android apps.
Each stream visits its instances three times, once per round, for 324, 411 and 120 episodes.
Each round is a seeded permutation of the instances, and the same stream file serves every run (\appref{app:streams}).
The environment is restored to its initial state at every round boundary on the web and before every task on mobile, so a later round cannot pass by inheriting the state an earlier one left behind.
The step budget is 30 on the web streams and 40 on \mw.

\par\noindent{\bf Agents.}
We adapt two open GUI agents, \uitars \citep{qin2025uitars} and \qwenvl \citep{qwen3vl2025}, from their released checkpoints with their own action spaces and no training before the stream.
Their prompts follow each release, with two notes shared by every method on the web streams, and are the benchmark's own on \mw (\appref{app:prompts}).
Each emits a thought and one action per step.

\par\noindent{\bf \method configuration.}
The adapter is a rank-16 LoRA on the output and MLP projections of the upper half of the decoder layers, zero-initialized.
The window holds $W{=}4$ episodes, the support is $K{=}4$, and each admission takes one AdamW step at learning rate $10^{-5}$.
The judge, the proposer and the verifier are all gpt-5-mini, used without fine-tuning; \tabref{tab:aux} swaps them for other models.
A relabeled prefix must hold at least two actions with a repeat rate of at most 0.3.
Prompts are given in \appref{app:prompts} and the remaining details in \appref{app:method}.

\par\noindent{\bf Baselines.}
The frozen agent is the same checkpoint run on the same stream without updates.
AWM-online \citep{wang2024awm} and Darwinian Memory (DMS) \citep{mi2026darwinian} are the two in-setting memory methods of \secref{sec:setting}, adapted from the authors' released code to our agents and harness, with the same judge as their success signal.
Both are training-free and keep what they learn in the prompt, so they differ from \method in what persists.
The deviations from the released code are listed in \appref{app:method}.

\par\noindent{\bf Metric.}
We report success rate over the stream under each benchmark's own evaluator.
Every method and variant is run three times, and the tables give the mean and standard deviation.
The runs behind every cell are listed in \appref{app:fullresults}.

\subsection{Main result: adaptation is feasible}
\label{sec:main}
\begin{table}[t]
  \caption{\textbf{Main result.} Success rate (\%) over the stream under the benchmark's own evaluator; mean $\pm$ std over three runs per cell. Rows per stream: \wa\ 324, \vwa\ 411, \mw\ 120. AWM-online and DMS are the in-setting memory baselines of \secref{sec:setting}, run on the identical streams.}
  \label{tab:main}
  \centering
  \small
  \setlength{\tabcolsep}{2pt}
  
\begin{tabular}{lcccccc}
  \toprule
   & \multicolumn{2}{c}{\textbf{\wa}} & \multicolumn{2}{c}{\textbf{\vwa}} & \multicolumn{2}{c}{\textbf{\mw}} \\
  \cmidrule(lr){2-3} \cmidrule(lr){4-5} \cmidrule(lr){6-7}
  \textbf{Method} & \textbf{\uitars} & \textbf{\qwenvl} & \textbf{\uitars} & \textbf{\qwenvl} & \textbf{\uitars} & \textbf{\qwenvl} \\
  \midrule
  Frozen agent & 20.8 $\pm$ 0.6 & 20.1 $\pm$ 0.8 & 16.7 $\pm$ 1.4 & 14.8 $\pm$ 0.6 & 6.1 $\pm$ 1.0 & 9.7 $\pm$ 1.0 \\
  AWM-online & 22.7 $\pm$ 1.2 & 18.7 $\pm$ 1.6 & 16.3 $\pm$ 1.3 & 14.7 $\pm$ 1.2 & \textbf{9.7 $\pm$ 1.0} & 10.8 $\pm$ 1.7 \\
  DMS & 22.4 $\pm$ 0.6 & 19.2 $\pm$ 0.5 & 17.0 $\pm$ 0.9 & 15.7 $\pm$ 0.9 & 7.5 $\pm$ 0.8 & 9.7 $\pm$ 0.5 \\
  \midrule
  \rowcolor{oursrow} \method (ours) & \textbf{25.8 $\pm$ 1.6} & \textbf{24.9 $\pm$ 0.9} & \textbf{20.0 $\pm$ 0.8} & \textbf{20.9 $\pm$ 1.9} & 9.2 $\pm$ 0.8 & \textbf{13.3 $\pm$ 2.2} \\
  \bottomrule
\end{tabular}

\end{table}
\tabref{tab:main} gives the main result.
On every stream and with both agents, \method's success rate exceeds the frozen agent's by three to six points, and in every cell the margin exceeds the sum of the two standard deviations.
The setting is therefore feasible as posed: an agent can improve its own weights during deployment from its own episodes, as read by auxiliary models, with no ground truth and one attempt per task occurrence.

The two memory baselines occupy the other admissible form of persistent state, and they read the same judge.
On the web streams they track the frozen agent within about two points in either direction, whereas \method exceeds the better of the two by three to six points with both agents.
On \mw, where the frozen agents complete fewer than one task in ten, the picture is closer: AWM-online matches \method with \uitars, at 9.7 against 9.2 with overlapping standard deviations, and trails it with \qwenvl, at 10.8 against 13.3, while DMS stays near the frozen agent with both.
On these streams, then, \method exceeds both memory methods on the web and is comparable to AWM-online on mobile.

\subsection{Ablations}
\label{sec:anatomy}
\begin{table}[t]
    \caption{\textbf{Ablations.} One component removed at a time from the full method, on the \qwenvl agent: the window, replaced by one step per admitted episode; \topk, replaced by one-hot imitation of the executed tokens; and \hind, leaving the success branch alone. Success rate (\%), mean $\pm$ std over three runs, and the mean over the two streams.}
  \label{tab:anatomy}
  \centering
  \small
  
\begin{tabular}{lccc}
  \toprule
  \textbf{Variant} & \textbf{\vwa} & \textbf{\wa} & \textbf{Mean} \\
  \midrule
  Frozen agent & 14.8 $\pm$ 0.6 & 20.1 $\pm$ 0.8 & 17.5 \\
  \midrule
  \rowcolor{oursrow} \method (full) & 20.9 $\pm$ 1.9 & 24.9 $\pm$ 0.9 & 22.9 \\
  w/o window (one step per episode) & 18.2 $\pm$ 1.2 & 23.0 $\pm$ 1.2 & 20.6 \\
  w/o \topk (one-hot targets) & 20.6 $\pm$ 2.6 & 22.0 $\pm$ 1.5 & 21.3 \\
  w/o \hind (success branch only) & 17.0 $\pm$ 1.4 & 21.9 $\pm$ 0.6 & 19.4 \\
  \bottomrule
\end{tabular}

\end{table}
\tabref{tab:anatomy} removes one component at a time from the full method on the \qwenvl agent and both web streams: every variant stays above the frozen agent, and every removal lowers the mean over the two streams.
\Hind matters most: the success branch alone keeps about a third of the margin, and it updates on about a quarter of the episodes on \vwa and a fifth on \wa, whereas the verified prefixes raise both shares to about half (\tabref{tab:ablfull}).
The window comes next: one step per admitted episode keeps a little over half of the margin, and its episodes stay as short as the full method's.
Replacing \topk with one-hot imitation of the executed tokens leaves \vwa level and costs three points on \wa.
The one-hot gradient is more than an order of magnitude larger at the same learning rate (\appref{app:fullresults}) and pushes every alternative toward zero, and our reading is that the policy over-commits to its recorded action sequences.
Its episodes lengthen round by round (\tabref{tab:eplen}): by the third round about three quarters of them on \wa run to the step budget, against one in five for the full method, and its success rate there falls back to the frozen level.
On \vwa the episodes lengthen in the same way but the wins hold within three rounds, so on these streams the target form matters only on \wa, whose episodes are longer.

\subsection{Sensitivity to the auxiliary models}
\label{sec:aux}
\begin{table}[t]
  \caption{\textbf{Dependence on the auxiliary models.} \method on the \qwenvl agent with the judge, the proposer and the verifier all replaced by the same alternative model. Success rate (\%), mean $\pm$ std over three runs, and the judge's precision, the share of the episodes it admitted that the benchmark evaluator scores as successes, averaged over the runs.}
  \label{tab:aux}
  \centering
  \small
  
\begin{tabular}{lcccc}
  \toprule
   & \multicolumn{2}{c}{\textbf{\vwa}} & \multicolumn{2}{c}{\textbf{\wa}} \\
  \cmidrule(lr){2-3} \cmidrule(lr){4-5}
  \textbf{Auxiliaries} & \textbf{Success (\%)} & \textbf{Judge prec.} & \textbf{Success (\%)} & \textbf{Judge prec.} \\
  \midrule
  Frozen agent (no auxiliaries) & 14.8 $\pm$ 0.6 & -- & 20.1 $\pm$ 0.8 & -- \\
  \midrule
  \rowcolor{oursrow} gpt-5-mini & 20.9 $\pm$ 1.9 & 0.51 & 24.9 $\pm$ 0.9 & 0.66 \\
  gpt-5 & 18.9 $\pm$ 1.4 & 0.59 & 23.0 $\pm$ 0.7 & 0.76 \\
  gpt-5-nano & 19.1 $\pm$ 0.6 & 0.43 & 23.9 $\pm$ 1.7 & 0.46 \\
  the agent's own model (\qwenvl) & 17.3 $\pm$ 0.8 & 0.35 & 24.2 $\pm$ 1.0 & 0.54 \\
  \bottomrule
\end{tabular}

\end{table}
\tabref{tab:aux} replaces the judge, the proposer and the verifier together with three alternatives to gpt-5-mini.
With gpt-5 or gpt-5-nano the success rate stays within two points on both streams, and every choice, including the agent's own 8B model, keeps the method above the frozen agent.
\method is therefore robust to the choice of auxiliary models.
The one row that separates is the agent's own model on \vwa, at 3.6 points below gpt-5-mini.
It is also the row with the lowest judge precision, about a third of the admitted episodes being true successes, whereas precision otherwise does not order the rows: gpt-5 is the most precise judge on both streams and does not give the highest success rate.
Our reading is that the judge acts as a soft filter rather than as a stand-in for the evaluator.
It admits the episodes that look like completions, and those carry a usable signal even where the evaluator would reject some of them, until precision falls far enough that the admitted set no longer resembles success.

\subsection{Where the gain appears}
\label{sec:mechanism}
\begin{figure}[t]
  \centering
  \includegraphics[width=\linewidth]{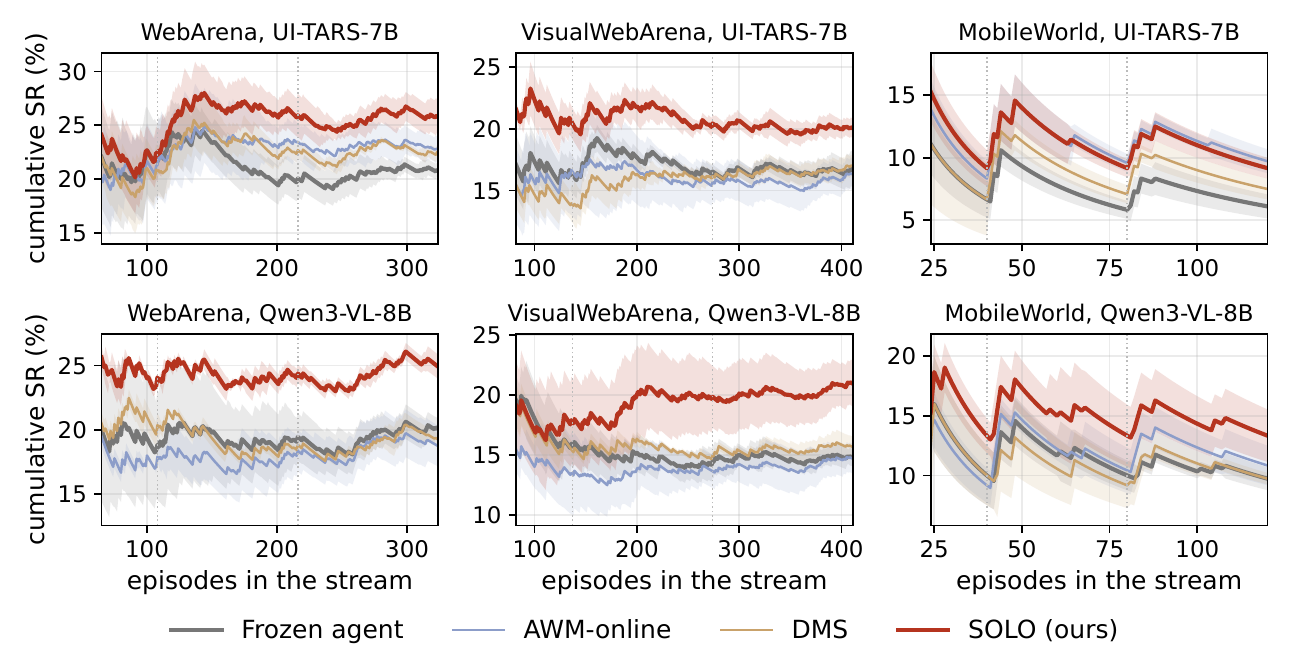}
  \caption{\textbf{Cumulative success rate along the stream.} Each panel is one cell of \tabref{tab:main}: the success rate over the episodes seen so far, averaged over the three runs, with a band of one standard deviation across the runs. Dotted lines mark the round boundaries. Each panel starts at twenty percent of its stream, where the earlier points rest on too few episodes.}
  \label{fig:cumulative}
\end{figure}
\figref{fig:cumulative} plots the cumulative success rate along each stream, and \tabref{tab:fullresults} gives the per-round counts behind it.
Read by round, the margin over the frozen agent takes two shapes.
With \qwenvl on \wa and \uitars on \vwa it is present from the first round, before any instance recurs, although templates already recur within that round (\appref{app:streams}).
With \uitars on \wa and \qwenvl on \vwa the first round ends within three wins of the frozen agent, and the margin opens in the second and third rounds, when the instances recur.
On \mw the per-round margins are one to two tasks of forty in every round and do not separate the two shapes.
We read this as adaptation to the deployment distribution rather than memorization of particular tasks.
As in fully test-time adaptation for perception models, where the model adjusts to the distribution of its test inputs, the agent here adjusts to the sites and errands of its stream, and the benefit shows in its online performance.

\section{Related work}
\label{sec:related}

\par\noindent{\bf Test-time adaptation.}
Test-time adaptation updates a trained model on unlabeled test inputs as they arrive \citep{liang2024ttasurvey}.
Tent minimizes prediction entropy and touches only normalization parameters \citep{wang2021tent}, test-time training adds a self-supervised head but must alter training to do so \citep{sun2020ttt}.
The continual variants guard against drift over a long stream, CoTTA by distilling toward a weight-averaged teacher and restoring weights to the source \citep{wang2022cotta}, EATA by selecting reliable samples and regularizing toward the source \citep{niu2022eata}.
\method relates to these at three points: its judge plays the role that confidence plays in EATA's sample selection, its \topk is the analogue of CoTTA's soft pseudo-labels, taken from the same forward pass rather than from a separate teacher, and its update of a small adapter parallels Tent's restriction to normalization layers.
The difference lies in the signal: a confident prediction is at once evidence and target, whereas an episode-level verdict names no action.

\par\noindent{\bf Adaptation through extra rollouts or a training phase.}
Language models are adapted at test time by fine-tuning on data derived from the test instance \citep{akyurek2024ttt} or by treating the majority vote over many samples as a reward \citep{zuo2025ttrl}.
For agents the extra compute goes into rollouts: GTA1 samples several candidate actions and lets a judge pick one \citep{yang2025gta1}, and JIT-RL, EvoTest and GTTA spend several attempts or exploration episodes on each task or environment \citep{li2026jitrl,he2025evotest,chen2026gtta_full}.
Online reinforcement learning spends a training phase instead, with an environment evaluator or a trained reward model, resets and parallel rollouts \citep{bai2024digirl,qi2025webrl,uitars2_2025,dreamgym2025,onlinerl_mobile2026}, and the flywheels behind self-evolving agents iterate that loop with reward models, synthesized tasks or rejection fine-tuning \citep{gao2025selfevolving,uigenie2025,lin2026uivoyager,magicgui_rms2026,agentevolver2025,jin2026sega}.
Each relies on a resource the setting withholds: samples or attempts beyond the one that counts, or a training phase with rewards.
\method also adapts from the agent's own experience, but from one attempt per occurrence and without a reward.

\par\noindent{\bf Experiential memory.}
Agents also improve by keeping experience in an external store.
Reflexion and ExpeL reflect on failed attempts and try again, or distill insights from a training split \citep{shinn2023reflexion,zhao2024expel}, MAGNET harvests trajectories over passes through the tasks before they count \citep{sun2026magnet}, and EchoTrail fills its memory in an exploration stage \citep{li2026echotrail}: these are practice-based and use rollouts that the one-attempt constraint withholds.
Stream-based systems learn as they go, admitting entries under a model judge or reflector, AWM-online by inducing workflows \citep{wang2024awm,pan2024autoeval}, Darwinian Memory by scoring and pruning \citep{mi2026darwinian}, ReasoningBank by distilling strategies \citep{reasoningbank2025}, and Mobile-Agent-E by evolving tips \citep{wang2025mobileagente}.
These operate in our setting with memory as the persistent state, the other admissible choice.
\method takes the weights instead, and \secref{sec:main} compares it with two of these systems on the same streams with the same judge.

\par\noindent{\bf Self-imitation and learning from failures.}
The objective relates to self-imitation and rejection sampling fine-tuning, which train on the agent's own outputs that pass a check \citep{oh2018sil,zelikman2022star,singh2024restem,yuan2023rft}, with LLM judges standing in for the check on GUI trajectories \citep{pan2024autoeval,uigenie2025}.
\method applies that selection online, one episode at a time, and distills toward the policy's own truncated distribution.
The failure branch relates to work on learning from failed trajectories: hindsight relabeling turns a failure into a success for the goal it reached \citep{andrychowicz2017her,zhang2023hir,agenther2026}, inverse-dynamics objectives and early experience learn from the transitions an agent's own actions produce, without a reward \citep{guishift2025,zhang2025earlyexperience}, and unlikelihood or preference losses train against the failed actions themselves \citep{welleck2020unlikelihood,ethayarajh2024kto}.
\method takes the first route: it relabels a verified prefix with the subtask it completed and admits it through the same update.

\section{Discussion and limitations}
\label{sec:discussion}
Within three rounds of a recurring stream, an off-the-shelf agent improves its own weights from its deployment episodes, read by auxiliary models, with no ground truth and one attempt per task occurrence, on three benchmarks and with two agents.
The claim concerns online performance on that stream: as with Tent, which adapts a model to the test distribution itself, the agent adapts to the sites and errands it is deployed on.
The streams are built from benchmark tasks that recur three times, not from user logs, so how often a real deployment repeats an errand, and hence the size of the gain there, is not measured.
On \mw the frozen agents pass fewer than one task in ten, so judged successes are rare.

The signal depends on auxiliary models: each episode costs one judge call and each judged failure at most two more.
With the agent's own 8B model in every auxiliary role the gain holds on \wa and shrinks on \vwa (\tabref{tab:aux}), and swapping either role alone accounts for most of that loss (\appref{app:judge}).
Our reading is that the judge acts as a soft filter: it reads the episode and not the environment, it admits what looks like a completion on screen, and such episodes carry a usable signal even when the evaluator would reject some of them, which is consistent with precision not ordering the rows of \tabref{tab:aux} and with the signal fading only when the admitted set stops resembling success.
The same channel is an attack surface: page content that sways the judge or the proposer trains the adapter, and we do not study adversarial pages.

Under the same judge, \method exceeded two prompt-memory methods on the web streams and was comparable to AWM-online on mobile (\tabref{tab:main}).
This is evidence for the in-weight route rather than a verdict between the two: memory keeps experience in an inspectable, deletable store and adapts in context, weights internalize it, and recent position papers argue that the two are complementary and that durable adaptation needs its own channel beside retrieval \citep{xu2026memo,dorovatas2026modular}.
The cost of persistence is drift.
By the third round the one-hot variant runs three quarters of its \wa episodes to the step budget, and the full method's episodes also lengthen with \qwenvl, by two to three steps with the share ending on the budget rising to about one in six, while with \uitars they shorten slightly (\tabref{tab:eplen}).
Behavior over longer deployments is untested, and the setting provides no ground truth with which to detect degradation.
Resetting the adapter to zero restores the frozen agent exactly, and detecting when to reset is not evaluated here.

\subsection*{AI use statement}
Large language and vision-language models appear in this work in two roles.
First, as components of the experiments: the judge, the proposer and the verifier of \method are gpt-5-mini by default and gpt-5, gpt-5-nano or \qwenvl in \secref{sec:aux}, the memory baselines use the same judge and, for AWM-online, the authors' gpt-4o induction model, and the VisualWebArena image-query evaluator is served by gpt-5-mini in place of BLIP-2.
Every such use is described in \secref{sec:setup} and \appref{app:method}, \appref{app:streams} and \appref{app:prompts}, with the prompts reproduced verbatim.
Second, as tools: AI coding assistants helped write and debug the experiment harness and the analysis scripts, and an AI writing assistant helped edit the text of this paper under the authors' direction.
All experiments were designed, run and checked by the authors, every number in the paper is a logged measurement from the runs listed in \appref{app:signal}, \appref{app:fullresults} and \appref{app:judge}, and the authors take full responsibility for the content.

\subsection*{Reproducibility statement}
The setting is defined in \secref{sec:setting} and the method in \secref{sec:method} and \algref{alg:solo}.
The configuration used in every run (adapter geometry, optimizer, window and support sizes, and the guards of the failure branch) is given in \secref{sec:setup} and \appref{app:method}, together with the deviations of the two adapted baselines from their released code.
The three streams are fixed files: their composition, order, resets and evaluators are described in \appref{app:streams}, and every run behind every table is listed in \appref{app:fullresults} with its per-round wins.
The prompts of the agents, the judge, the proposer, the verifier and the baselines are reproduced verbatim in \appref{app:prompts}.
The benchmarks are the public WebArena, VisualWebArena and MobileWorld releases, self-hosted.
We will release the code, the stream files and the run logs behind every number.

\subsection*{Ethics statement}
All experiments ran on self-hosted copies of the benchmark websites and on Android emulators.
No real user account, no real person's data and no live service was involved, and the screenshots sent to the auxiliary models show only these sandboxed environments.
The setting studied here, an agent that updates its own weights while deployed, carries risks beyond those of a frozen agent.
A GUI agent's actions can be irreversible, an agent that learns from what it sees is exposed to page content crafted to sway its judge or its proposer, and a self-updating policy can drift without any ground truth to detect it.
\secref{sec:discussion} states these risks and the one safeguard the design leaves available, resetting the adapter, which restores the frozen agent exactly.
We do not study adversarial pages, and we regard such a study as a prerequisite for any deployment of the method outside a sandbox.

\bibliography{refs}
\bibliographystyle{iclr2027_conference}

\newpage
\appendix

\section{Signal study: entropy minimization}
\label{app:signal}
The canonical unsupervised signal of test-time adaptation for perception models is the model's own confidence: Tent adapts by minimizing the entropy of its predictions \citep{wang2021tent}.
\tabref{tab:entmin} applies the same signal to a GUI agent.
This variant keeps \method's adapter, window and step size and replaces the loss with the entropy of the agent's next-token distribution, averaged over the positions of its own response, applied to every episode, with no judge and no relabeling.
The agent is \qwenvl on both web streams, three runs each.

Entropy minimization does not help on either stream, and on \vwa it collapses the agent.
The response entropy falls by an order of magnitude along the stream in every run.
On \vwa the policy stops acting: episodes that end at their first step, with an answer or a declaration of completion, rise from about a quarter of a round under the frozen agent to between half and most of the third round, and the wins fall from the frozen level in the first round to a handful in the third.
On \wa the entropy falls in the same way, but the agent keeps acting and its success rate stays at the frozen level.
Sharpening the policy toward what it already prefers therefore either collapses it or leaves it where it was, and the signal of \method comes from a judge instead.
\begin{table}[h]
  \caption{\textbf{Entropy minimization as the test-time signal.} \qwenvl, both web streams, three runs each, against the frozen agent. Wins per round with the total, episodes ended at their first step per round, and the mean per-episode response entropy in nats over the first quarter of the stream and over the last.}
  \label{tab:entmin}
  \centering
  \scriptsize
  \setlength{\tabcolsep}{3pt}
  \begin{tabular}{llccc}
  \toprule
  \textbf{Stream} & \textbf{Arm} & \textbf{Wins per round (total)} & \textbf{One-step episodes} & \textbf{Entropy} \\
  \midrule
  \vwa & frozen (mean of 3 runs) & 22 / 18 / 21 (61) & 36 / 35 / 33 & -- \\
   & entropy min., run 1 & 18 / 4 / 1 (23) & 55 / 90 / 117 & 0.116 $\to$ 0.014 \\
   & entropy min., run 2 & 22 / 10 / 4 (36) & 48 / 53 / 83 & 0.113 $\to$ 0.012 \\
   & entropy min., run 3 & 20 / 6 / 8 (34) & 48 / 87 / 76 & 0.114 $\to$ 0.014 \\
  \midrule
  \wa & frozen (mean of 3 runs) & 20 / 22 / 23 (65) & 5 / 6 / 6 & -- \\
   & entropy min., run 1 & 21 / 24 / 21 (66) & 6 / 2 / 2 & 0.111 $\to$ 0.016 \\
   & entropy min., run 2 & 22 / 19 / 17 (58) & 3 / 3 / 5 & 0.107 $\to$ 0.015 \\
   & entropy min., run 3 & 24 / 24 / 22 (70) & 6 / 8 / 18 & 0.113 $\to$ 0.014 \\
  \bottomrule
\end{tabular}

\end{table}

\section{Method details}
\label{app:method}
\paragraph{Objective and gradient.}
For a position $t$ with logits $z_t$, the gradient of \eqref{eq:topk} is $\partial\mathcal{L}_K/\partial z_t = p_\theta(\cdot\mid\xi_{<t}) - \tilde q_t$, where $\tilde q_t$ equals $q_t$ on the support $\mathcal{V}_K(t)$ and zero off it.
At the pre-update point, with $S_t=\sum_{v\in\mathcal{V}_K(t)}p_{\bar\theta}(v\mid\xi_{<t})$ the mass of the support, the component on a support token $v$ is $-p_{\bar\theta}(v)(1-S_t)/S_t$ and the component on any other token is $p_{\bar\theta}(v)$, so the gradient moves exactly the tail mass $1-S_t$ from outside the support onto it, in proportion to the current probabilities, and vanishes where the support already holds all the mass.
The targets come from the same forward pass with gradients stopped.
The supervised positions are all response tokens of every step of the episode, reasoning and action alike.
Each step is conditioned on the instruction, the screenshots seen so far and the earlier responses, exactly as it was served, and a \vwa task's input images precede the screenshots.
The loss is summed over an episode's supervised positions and divided by their number, then averaged over the episodes in the window, and one optimizer step follows.
A relabeled prefix is trained with the same scope under its own instruction.

\paragraph{Adapter and optimizer.}
The adapter is a LoRA of rank 16 and scale $\alpha=32$ without dropout on the attention output projection and the three MLP projections of the upper half of the decoder layers of the language model, with the vision encoder untouched.
Its $A$ matrices are drawn from $\mathcal{N}(0,0.02)$ with a fixed seed and its $B$ matrices are zero, so the adapted policy starts identical to the frozen agent.
The optimizer is AdamW at learning rate $10^{-5}$ without weight decay, the gradient norm is clipped at one, and the model runs in bfloat16.
The window holds $W=4$ episodes and the support is $K=4$.
Every variant of \tabref{tab:anatomy} and \tabref{tab:aux} keeps these settings and changes only the named component, and both agents use the same values on every stream.

\paragraph{Failure branch.}
The proposer sees the original instruction and, for every step, the screenshot the agent saw, its reasoning and its action.
It answers whether some prefix completed a subtask, prerequisite or narrower version of the task, the earliest step $k$ at which that was completed, and the instruction $g'$ that names it, or abstains.
Before verification the prefix up to $k$, with a trailing declaration of completion dropped, must hold at least two actions, and the share of its steps that repeat the previous action must not exceed 0.3.
The verifier sees only $g'$, the actions of the prefix and the screenshot after step $k$, and answers whether that screenshot shows $g'$ completed.
An episode that ended at its first step has no prefix and is skipped.
An admitted prefix enters the window under $g'$ with the same weight and the same loss scope as a judged success, and the anchor holds any update until the window holds a judged success.
\tabref{tab:hind} gives the fate of the judged failures in the three runs of each cell.
The guards are the largest filter, and most of what they reject is a proposed prefix of a single action.
The verifier rejects between one in thirteen and one in four of the prefixes that reach it.
The admitted prefixes hold three to four actions on average.
On \wa about half of the failures end as admitted prefixes, on \mw about three fifths, and on \vwa about a quarter.
The anchor withheld an update at about one admission in seven for \qwenvl on \wa, at about one in twenty for \uitars on both web streams, and never on \mw or for \qwenvl on \vwa.
\begin{table}[h]
  \caption{\textbf{What becomes of a judged failure.} Per stream and agent, mean over the three \method runs: episodes, judged successes, judged failures, and of the failures those with no prefix (ended at the first step), those on which the proposer abstained, those rejected by the guards (in parentheses, rejected as a single-action prefix), those rejected by the verifier, and those admitted as relabeled prefixes, with the admitted share of the failures and the mean number of steps in an admitted prefix.}
  \label{tab:hind}
  \centering
  \scriptsize
  \setlength{\tabcolsep}{3pt}
  \begin{tabular}{llrrrrrrrrr}
  \toprule
  \textbf{Stream} & \textbf{Agent} & \textbf{Episodes} & \textbf{Succ.} & \textbf{Fail.} & \textbf{No prefix} & \textbf{Abstain} & \textbf{Guards} & \textbf{Verifier} & \textbf{Admitted} & \textbf{Prefix steps} \\
  \midrule
  \wa & \uitars & 324 & 96 & 228 & 1 & 42 & 63 (53) & 22 & 101 (44\%) & 3.7 \\
  \wa & \qwenvl & 324 & 74 & 250 & 6 & 23 & 69 (51) & 19 & 134 (54\%) & 3.0 \\
  \midrule
  \vwa & \uitars & 411 & 94 & 317 & 2 & 67 & 133 (100) & 24 & 91 (29\%) & 3.3 \\
  \vwa & \qwenvl & 411 & 127 & 284 & 49 & 45 & 100 (83) & 20 & 69 (24\%) & 3.0 \\
  \midrule
  \mw & \uitars & 120 & 23 & 97 & 0 & 12 & 19 (13) & 10 & 56 (58\%) & 3.9 \\
  \mw & \qwenvl & 120 & 25 & 95 & 0 & 6 & 22 (16) & 5 & 62 (65\%) & 3.6 \\
  \bottomrule
\end{tabular}

\end{table}

\paragraph{Baselines.}
AWM-online follows the online loop of the authors' released code \citep{wang2024awm}: after every judged success the workflow file of the site is re-induced from scratch over all judged-positive episodes so far, deduplicated by template with one example per template, with the authors' induction prompt verbatim, at temperature one, and the completion is the workflow file that the agent then sees in its prompt.
Three deviations adapt the method to our agents and harness.
The one-shot example and the appended action-space note are rewritten in our agents' action vocabulary, since the original agent acts on accessibility trees.
The induction model stays the authors' gpt-4o, because gpt-5-mini returned empty inductions at the authors' token budget, while the judge that gates memory writes is gpt-5-mini as for \method.
The rendered workflow block is capped at four thousand characters, since a 7B context cannot hold an unbounded library.
Darwinian Memory follows the authors' released code \citep{mi2026darwinian}, an anonymized repository linked from the paper that we accessed on 18 September 2026 and that has since expired: their memory score with their constants and logical clock, their strike semantics with deletion at three strikes, and their retrieval with rerank and top three, verified by the same judge.
Darwinian Memory needs two deviations.
The similarity is an IDF-weighted token cosine instead of a sentence-embedding model, with the admission threshold recalibrated from 0.8 to 0.15, which admits no cross-template pair and keeps 64\% of same-template pairs on the \wa intents.
Their planner reputation manager regulates a planner's plans and has no counterpart in a single-screenshot policy.
Both baselines keep the policy frozen and everything they learn in the prompt.

\section{Streams and protocol}
\label{app:streams}
\paragraph{\wa.}
The stream holds 108 instances of 27 WebArena templates, four per template: ten templates on the shopping site, nine on GitLab, four on the shopping admin panel and four on the forum.

\paragraph{\vwa.}
The stream holds 137 instances of 39 VisualWebArena templates, three or four per template: 28 templates on the shopping site, nine on classifieds and two on the forum.

\paragraph{\mw.}
MobileWorld tasks have no parameterized instances, so each task is one instance.
The stream holds 40 tasks: sixteen on Mastodon, nine on Gmail, eight on native Android apps and seven on device settings.

\paragraph{Order and resets.}
Every stream has three rounds.
Within a round the instances are permuted with a seed, and the permutation is redrawn until no instance recurs within eight rows of its previous visit, so the same order serves every run.
On the web streams the sites are restored to their initial state at every round boundary through the benchmarks' reset procedures, since many templates change the state of a site and a later round would otherwise inherit what an earlier one left behind.
On \mw the benchmark's own setup and teardown restore the device before every task.
The step budget is 30 on the web streams and 40, the official budget, on \mw.

\paragraph{Evaluators.}
The evaluators are the benchmarks' own and are never visible to the agent or to the auxiliary models.
VisualWebArena's image-query evaluator takes a visual question answering function, which gpt-5-mini serves in place of the reference implementation's BLIP-2.
MobileWorld's tasks are graded by the benchmark's checkers.

\section{Full results}
\label{app:fullresults}
\tabref{tab:fullresults} lists the three runs behind every cell of \tabref{tab:main}: total wins with the per-round split, the mean and std of wins, and the success rate.
\tabref{tab:ablfull} lists the runs behind every row of \tabref{tab:anatomy} in the same form, and \tabref{tab:auxfull} lists every run behind \tabref{tab:aux} with the judge's precision on each.
\begin{table}[h]
  \caption{\textbf{Every run behind \tabref{tab:main}.} Wins per run with the round split in parentheses, then mean $\pm$ std of wins and success rate (\%).}
  \label{tab:fullresults}
  \centering
  \scriptsize
  \setlength{\tabcolsep}{3pt}
  
\begin{tabular}{lllcccrr}
  \toprule
  \textbf{Benchmark} & \textbf{Method} & \textbf{Agent} & \textbf{Run 1} & \textbf{Run 2} & \textbf{Run 3} & \textbf{Mean $\pm$ std} & \textbf{SR (\%)} \\
  \midrule
  \wa & Frozen agent & \uitars & 68 (26/19/23) & 69 (25/21/23) & 65 (19/20/26) & 67.33 $\pm$ 2.08 & 20.8 \\
   & Frozen agent & \qwenvl & 63 (15/23/25) & 64 (23/19/22) & 68 (23/23/22) & 65.00 $\pm$ 2.65 & 20.1 \\
   & AWM-online & \uitars & 72 (23/28/21) & 78 (27/27/24) & 71 (19/28/24) & 73.67 $\pm$ 3.79 & 22.7 \\
   & AWM-online & \qwenvl & 62 (19/21/22) & 55 (19/17/19) & 65 (20/23/22) & 60.67 $\pm$ 5.13 & 18.7 \\
   & DMS & \uitars & 72 (22/28/22) & 71 (25/25/21) & 75 (20/27/28) & 72.67 $\pm$ 2.08 & 22.4 \\
   & DMS & \qwenvl & 62 (21/20/21) & 61 (22/18/21) & 64 (23/18/23) & 62.33 $\pm$ 1.53 & 19.2 \\
  \rowcolor{oursrow} & \method (ours) & \uitars & 89 (25/34/30) & 79 (21/30/28) & 83 (26/32/25) & 83.67 $\pm$ 5.03 & 25.8 \\
  \rowcolor{oursrow} & \method (ours) & \qwenvl & 80 (26/27/27) & 78 (26/25/27) & 84 (26/28/30) & 80.67 $\pm$ 3.06 & 24.9 \\
  \midrule
  \vwa & Frozen agent & \uitars & 62 (23/21/18) & 73 (24/23/26) & 71 (20/25/26) & 68.67 $\pm$ 5.86 & 16.7 \\
   & Frozen agent & \qwenvl & 64 (24/20/20) & 59 (19/18/22) & 60 (22/16/22) & 61.00 $\pm$ 2.65 & 14.8 \\
   & AWM-online & \uitars & 63 (25/16/22) & 73 (26/22/25) & 65 (17/23/25) & 67.00 $\pm$ 5.29 & 16.3 \\
   & AWM-online & \qwenvl & 55 (13/20/22) & 61 (21/18/22) & 65 (21/19/25) & 60.33 $\pm$ 5.03 & 14.7 \\
   & DMS & \uitars & 66 (17/26/23) & 73 (21/24/28) & 70 (19/26/25) & 69.67 $\pm$ 3.51 & 17.0 \\
   & DMS & \qwenvl & 64 (22/19/23) & 68 (21/21/26) & 61 (20/20/21) & 64.33 $\pm$ 3.51 & 15.7 \\
  \rowcolor{oursrow} & \method (ours) & \uitars & 86 (29/29/28) & 81 (29/27/25) & 80 (26/28/26) & 82.33 $\pm$ 3.21 & 20.0 \\
  \rowcolor{oursrow} & \method (ours) & \qwenvl & 82 (22/31/29) & 95 (30/32/33) & 81 (21/27/33) & 86.00 $\pm$ 7.81 & 20.9 \\
  \midrule
  \mw & Frozen agent & \uitars & 6 (2/2/2) & 8 (3/2/3) & 8 (3/2/3) & 7.33 $\pm$ 1.15 & 6.1 \\
   & Frozen agent & \qwenvl & 11 (3/4/4) & 13 (4/4/5) & 11 (5/4/2) & 11.67 $\pm$ 1.15 & 9.7 \\
   & AWM-online & \uitars & 11 (3/5/3) & 13 (4/4/5) & 11 (3/4/4) & 11.67 $\pm$ 1.15 & 9.7 \\
   & AWM-online & \qwenvl & 15 (4/6/5) & 13 (4/4/5) & 11 (3/4/4) & 13.00 $\pm$ 2.00 & 10.8 \\
   & DMS & \uitars & 10 (2/4/4) & 9 (4/2/3) & 8 (2/3/3) & 9.00 $\pm$ 1.00 & 7.5 \\
   & DMS & \qwenvl & 11 (3/3/5) & 12 (5/4/3) & 12 (4/3/5) & 11.67 $\pm$ 0.58 & 9.7 \\
  \rowcolor{oursrow} & \method (ours) & \uitars & 12 (4/4/4) & 10 (3/4/3) & 11 (4/3/4) & 11.00 $\pm$ 1.00 & 9.2 \\
  \rowcolor{oursrow} & \method (ours) & \qwenvl & 19 (6/7/6) & 15 (5/5/5) & 14 (5/4/5) & 16.00 $\pm$ 2.65 & 13.3 \\
  \bottomrule
\end{tabular}

\end{table}
\begin{table}[h]
  \caption{\textbf{Every run behind \tabref{tab:anatomy}.} \qwenvl on both web streams. Wins per run with the round split in parentheses, then mean $\pm$ std of wins, success rate (\%), and the mean number of updates per run.}
  \label{tab:ablfull}
  \centering
  \scriptsize
  \setlength{\tabcolsep}{3pt}
  \begin{tabular}{llcccrrr}
  \toprule
  \textbf{Stream} & \textbf{Variant} & \textbf{Run 1} & \textbf{Run 2} & \textbf{Run 3} & \textbf{Mean $\pm$ std} & \textbf{SR (\%)} & \textbf{Updates} \\
  \midrule
  \vwa & Frozen agent & 64 (24/20/20) & 59 (19/18/22) & 60 (22/16/22) & 61.00 $\pm$ 2.65 & 14.8 & -- \\
  \cmidrule(l){2-8}
  \rowcolor{oursrow} & \method (full) & 82 (22/31/29) & 95 (30/32/33) & 81 (21/27/33) & 86.00 $\pm$ 7.81 & 20.9 & 196 \\
   & w/o window & 74 (20/28/26) & 80 (23/29/28) & 70 (20/25/25) & 74.67 $\pm$ 5.03 & 18.2 & 176 \\
   & w/o \topk & 75 (19/28/28) & 83 (21/30/32) & 96 (26/37/33) & 84.67 $\pm$ 10.60 & 20.6 & 180 \\
   & w/o \hind & 64 (19/18/27) & 71 (24/22/25) & 75 (24/25/26) & 70.00 $\pm$ 5.57 & 17.0 & 109 \\
  \midrule
  \wa & Frozen agent & 63 (15/23/25) & 64 (23/19/22) & 68 (23/23/22) & 65.00 $\pm$ 2.65 & 20.1 & -- \\
  \cmidrule(l){2-8}
  \rowcolor{oursrow} & \method (full) & 80 (26/27/27) & 78 (26/25/27) & 84 (26/28/30) & 80.67 $\pm$ 3.06 & 24.9 & 177 \\
   & w/o window & 79 (23/28/28) & 73 (22/29/22) & 72 (20/29/23) & 74.67 $\pm$ 3.79 & 23.0 & 196 \\
   & w/o \topk & 68 (18/27/23) & 77 (22/34/21) & 69 (24/26/19) & 71.33 $\pm$ 4.93 & 22.0 & 180 \\
   & w/o \hind & 71 (20/26/25) & 69 (20/26/23) & 73 (21/27/25) & 71.00 $\pm$ 2.00 & 21.9 & 64 \\
  \bottomrule
\end{tabular}

\end{table}
\begin{table}[h]
  \caption{\textbf{Every run behind \tabref{tab:aux}.} Wins per run, the judge's precision on that run (share of admitted episodes that the evaluator scores as successes), and the success rate over the runs.}
  \label{tab:auxfull}
  \centering
  \scriptsize
  \setlength{\tabcolsep}{3pt}
  
\begin{tabular}{llccr}
  \toprule
  \textbf{Stream} & \textbf{Auxiliaries} & \textbf{Wins per run} & \textbf{Judge precision per run} & \textbf{SR (\%)} \\
  \midrule
  \rowcolor{oursrow} \vwa & gpt-5-mini & 82 / 95 / 81 & 0.52 / 0.54 / 0.48 & 20.9 \\
  \vwa & gpt-5 & 76 / 73 / 84 & 0.59 / 0.59 / 0.60 & 18.9 \\
  \vwa & gpt-5-nano & 80 / 76 / 80 & 0.46 / 0.41 / 0.42 & 19.1 \\
  \vwa & the agent's own model (\qwenvl) & 69 / 75 / 69 & 0.36 / 0.36 / 0.34 & 17.3 \\
  \midrule
  \rowcolor{oursrow} \wa & gpt-5-mini & 80 / 78 / 84 & 0.63 / 0.67 / 0.67 & 24.9 \\
  \wa & gpt-5 & 72 / 76 / 76 & 0.74 / 0.75 / 0.78 & 23.0 \\
  \wa & gpt-5-nano & 80 / 71 / 81 & 0.49 / 0.44 / 0.46 & 23.9 \\
  \wa & the agent's own model (\qwenvl) & 82 / 77 / 76 & 0.53 / 0.50 / 0.58 & 24.2 \\
  \bottomrule
\end{tabular}

\end{table}

\paragraph{Runs and variance.}
Every method and variant is run three times on the identical stream in the identical order.
Runs differ through the nondeterminism of the environments, of the agents' inference and of the auxiliary models.
Every table reports the mean and the sample standard deviation of wins over the three runs, and the success rate of the mean.
\figref{fig:cumulative} is built from the same runs: for each run the success rate over the episodes seen so far, then the mean over the three runs of a cell with a band of one standard deviation across them.

\paragraph{Episode length.}
\tabref{tab:eplen} gives the mean number of steps per episode and the share of episodes that end on the step budget, per round, for every cell of \tabref{tab:main} on the web streams and every variant of \tabref{tab:anatomy}.
The frozen agents and the memory baselines are flat across rounds.
With \qwenvl the full method's episodes lengthen over the three rounds, by two to three steps, and the share ending on the budget rises to about one in six.
With \uitars its episodes shorten slightly and fewer end on the budget than for the frozen agent.
The one-hot variant lengthens its episodes far more on both streams.
Its updates are also larger: the median gradient norm of a one-hot update is 0.56 to 0.61 in the three runs of each stream, against 0.02 for \topk at the same learning rate.
\begin{table}[h]
  \caption{\textbf{Episode length along the stream.} Mean steps per episode and share of episodes ending on the step budget, per round, mean over the three runs. Every cell of \tabref{tab:main} on the web streams and every variant of \tabref{tab:anatomy}.}
  \label{tab:eplen}
  \centering
  \scriptsize
  \setlength{\tabcolsep}{3pt}
  \begin{tabular}{lllcc}
  \toprule
  \textbf{Stream} & \textbf{Agent} & \textbf{Method} & \textbf{Steps per episode (r1 / r2 / r3)} & \textbf{Budget-ended \% (r1 / r2 / r3)} \\
  \midrule
  \wa & \uitars & frozen agent & 17.0 / 16.5 / 17.5 & 41 / 40 / 43 \\
   &  & AWM-online & 17.4 / 16.4 / 17.5 & 42 / 36 / 40 \\
   &  & DMS & 16.7 / 16.0 / 17.5 & 41 / 36 / 43 \\
  \rowcolor{oursrow} &  & \method & 17.8 / 15.7 / 16.1 & 43 / 34 / 35 \\
  \midrule
   & \qwenvl & frozen agent & 9.1 / 9.5 / 9.1 & 6 / 8 / 8 \\
   &  & AWM-online & 9.0 / 8.1 / 8.0 & 9 / 7 / 5 \\
   &  & DMS & 9.1 / 9.0 / 8.9 & 10 / 7 / 8 \\
  \rowcolor{oursrow} &  & \method & 9.4 / 11.0 / 11.9 & 7 / 12 / 19 \\
   &  & w/o window & 9.5 / 10.8 / 11.9 & 10 / 14 / 17 \\
   &  & w/o \topk & 9.3 / 17.3 / 25.0 & 8 / 36 / 74 \\
   &  & w/o \hind & 9.1 / 9.6 / 9.9 & 6 / 9 / 10 \\
  \midrule
  \vwa & \uitars & frozen agent & 15.4 / 16.4 / 16.5 & 40 / 43 / 44 \\
   &  & AWM-online & 16.3 / 15.9 / 16.8 & 42 / 40 / 45 \\
   &  & DMS & 16.3 / 16.5 / 15.8 & 43 / 44 / 41 \\
  \rowcolor{oursrow} &  & \method & 15.8 / 14.2 / 14.6 & 41 / 35 / 36 \\
  \midrule
   & \qwenvl & frozen agent & 4.8 / 4.9 / 5.4 & 2 / 1 / 3 \\
   &  & AWM-online & 5.2 / 4.7 / 5.2 & 2 / 2 / 2 \\
   &  & DMS & 5.0 / 4.7 / 4.7 & 3 / 2 / 1 \\
  \rowcolor{oursrow} &  & \method & 6.0 / 7.2 / 8.9 & 4 / 9 / 15 \\
   &  & w/o window & 5.2 / 6.2 / 7.9 & 3 / 5 / 10 \\
   &  & w/o \topk & 5.2 / 11.0 / 16.3 & 3 / 19 / 42 \\
   &  & w/o \hind & 5.2 / 6.0 / 7.9 & 2 / 6 / 10 \\
  \bottomrule
\end{tabular}

\end{table}

\section{Judge and auxiliary models}
\label{app:judge}
\paragraph{The judge against the evaluator.}
\tabref{tab:judge} measures the gpt-5-mini judge on \method's own episodes in the three runs of each cell: how many episodes it admitted, how many of those the benchmark evaluator scores as successes, and how many evaluator successes there were.
On the web streams the judge admits between six and eight of every ten evaluator successes, and between a third and a half of what it admits is not an evaluator success.
On \mw it admits nearly every evaluator success, plus about as many episodes that the evaluator rejects with \uitars and about half as many with \qwenvl.
\begin{table}[h]
  \caption{\textbf{The gpt-5-mini judge on \method's episodes.} Mean over the three runs per cell: judged successes, the true successes among them, the evaluator's successes, precision (true among judged) and recall (judged among true).}
  \label{tab:judge}
  \centering
  \footnotesize
  \setlength{\tabcolsep}{3pt}
  \begin{tabular}{llrrrcc}
  \toprule
  \textbf{Stream} & \textbf{Agent} & \textbf{Judged} & \textbf{True among judged} & \textbf{Evaluator} & \textbf{Precision} & \textbf{Recall} \\
  \midrule
  \wa & \uitars & 96 & 58 & 84 & 0.60 & 0.69 \\
  \wa & \qwenvl & 74 & 49 & 81 & 0.66 & 0.60 \\
  \midrule
  \vwa & \uitars & 94 & 57 & 82 & 0.61 & 0.70 \\
  \vwa & \qwenvl & 127 & 65 & 86 & 0.51 & 0.76 \\
  \midrule
  \mw & \uitars & 23 & 10 & 11 & 0.44 & 0.94 \\
  \mw & \qwenvl & 25 & 16 & 16 & 0.63 & 0.98 \\
  \bottomrule
\end{tabular}

\end{table}

\paragraph{A perfect judge.}
\tabref{tab:gtjudge} replaces the judge by the benchmark's own evaluator, with gpt-5-mini kept as the proposer and the verifier, so that every admitted episode is a true success.
A perfect judge gives the same success rate as gpt-5-mini on both streams, within the run-to-run spread, although gpt-5-mini's admissions are a third to a half false.
Removing every false admission therefore does not raise the success rate beyond the run-to-run spread.
This is consistent with the soft-filter reading of \secref{sec:aux}.
\begin{table}[h]
  \caption{\textbf{A perfect judge against the default judge.} \method on \qwenvl with the judge replaced by the benchmark evaluator, proposer and verifier unchanged. Wins per run, judge precision, and success rate (\%) mean $\pm$ std over three runs.}
  \label{tab:gtjudge}
  \centering
  \footnotesize
  \setlength{\tabcolsep}{3pt}
  
\begin{tabular}{llccc}
  \toprule
  \textbf{Stream} & \textbf{Judge} & \textbf{Wins per run} & \textbf{Judge precision} & \textbf{Success (\%)} \\
  \midrule
  \rowcolor{oursrow} \vwa & gpt-5-mini (default) & 82 / 95 / 81 & 0.51 & 20.9 $\pm$ 1.9 \\
  \vwa & benchmark evaluator & 80 / 78 / 95 & 1.00 & 20.5 $\pm$ 2.3 \\
  \midrule
  \rowcolor{oursrow} \wa & gpt-5-mini (default) & 80 / 78 / 84 & 0.66 & 24.9 $\pm$ 0.9 \\
  \wa & benchmark evaluator & 78 / 71 / 78 & 1.00 & 23.4 $\pm$ 1.2 \\
  \bottomrule
\end{tabular}

\end{table}

\paragraph{The two roles swapped separately.}
\tabref{tab:split} swaps the two auxiliary roles one at a time on \vwa, the stream on which the agent's own model loses the most in \tabref{tab:aux}.
With the agent's own model as the judge alone, precision falls to about a third and the success rate to 18.2.
With it as the proposer and the verifier alone, precision is unchanged but the verified prefixes fall by about half, from about seventy to about thirty-five per run, and the success rate to 17.4.
With both roles swapped the success rate is 17.3.
Either role alone thus accounts for most of the loss, and the two together add little more.
\begin{table}[h]
  \caption{\textbf{The auxiliary roles swapped one at a time on \vwa.} \method on \qwenvl. Wins per run, judge precision and success rate (\%) mean $\pm$ std over three runs.}
  \label{tab:split}
  \centering
  \footnotesize
  \setlength{\tabcolsep}{3pt}
  
\begin{tabular}{lllcc}
  \toprule
  \textbf{Judge} & \textbf{Proposer and verifier} & \textbf{Wins per run} & \textbf{Judge precision} & \textbf{Success (\%)} \\
  \midrule
  \rowcolor{oursrow} gpt-5-mini & gpt-5-mini & 82 / 95 / 81 & 0.51 & 20.9 $\pm$ 1.9 \\
  \qwenvl & gpt-5-mini & 71 / 70 / 83 & 0.36 & 18.2 $\pm$ 1.8 \\
  gpt-5-mini & \qwenvl & 76 / 69 / 70 & 0.48 & 17.4 $\pm$ 0.9 \\
  \qwenvl & \qwenvl & 69 / 75 / 69 & 0.35 & 17.3 $\pm$ 0.8 \\
  \bottomrule
\end{tabular}

\end{table}

\section{Compute and cost}
\label{app:cost}
\method adds to each episode one judge call, a proposer and a verifier call on most judged failures, and one gradient step over the window.
In the reported runs the auxiliary calls number about two per episode.
A proposer call carries every screenshot of the failed episode and runs to five to seven thousand tokens, and a verifier call to about one thousand.
On the web streams these additions took between about half a minute and a minute per episode, measured within each run as the time from the end of the rollout to the end of the episode, against rollouts of one to one and a half minutes.
Every run occupied one NVIDIA H200 shared with other runs, so absolute times vary with the load at the time of the run and are not compared across runs.

\section{Prompts}
\label{app:prompts}
Every prompt below is reproduced as used in the reported runs, folded to plain ASCII and re-wrapped to the page width.
Braces mark the fields filled at run time.

\paragraph{Agents.}
\uitars runs with its own web prompt, followed by two notes that every method on the web streams shares, one on information tasks and one on step discipline.
\qwenvl runs with the computer-use tool prompt of its release and the response format of the MobileWorld agent, and the same two notes ride inside the user turn after the instruction.
On \mw both agents use the benchmark's own prompts, \uitars with the information-task note.
{\footnotesize
\begin{verbatim}
You are a GUI agent. You are given a task and your action history,
  with screenshots. You need to perform the next action to complete
  the task.

## Output Format
```
Thought: ...
Action: ...
```

## Action Space
click(start_box='<|box_start|>(x1,y1)<|box_end|>')
type(content='') # If you want to submit your input, use "\n" at the
  end of `content`.
scroll(start_box='<|box_start|>(x1,y1)<|box_end|>',
  end_box='<|box_start|>(x3,y3)<|box_end|>')
press_back() # Go back to the previous page.
press_home() # Go back to the task's starting page.
finished(content='') # Submit the task regardless of whether it
  succeeds or fails.
answer(content='') # Answer user's question.

## Note
- Use English in `Thought` and `Action` part.
- Write a small plan and finally summarize your next action (with its
  target element) in one sentence in `Thought` part.

## User Instruction
{instruction}


## Information tasks
For information-seeking tasks (questions asking how many / how long /
  what / which / to answer with ...), return the requested information
  using answer(content='...') -- do NOT terminate with an empty
  finished(). Only call finished() for non-question tasks.

## Step discipline
Your step budget is limited. Before each action, check whether the
  task goal is already met on the current page -- if it is, emit
  finished() (or answer(content='...') for questions) immediately
  instead of acting further. If your previous action did not change
  the page, do not repeat it -- choose a different element or a
  different action type.
\end{verbatim}}
{\footnotesize
\begin{verbatim}
You are a helpful assistant.

# Tools

You may call one or more functions to assist with the user query.

You are provided with function signatures within <tools></tools> XML
  tags:
<tools>
{"type": "function", "function": {"name": "computer_use",
  "description": "Use a mouse and keyboard to interact with a
  computer, and take screenshots.\n* This is an interface to a web
  page in a browser. The screenshot shows the page content only (no
  address bar or browser buttons); use the `back` and `home` actions
  to navigate between pages.\n* Some pages may take time to load or
  process actions, so you may need to wait and take successive
  screenshots to see the results of your actions.\n* The screen's
  resolution is 1000x1000.\n* Whenever you intend to move the cursor
  to click on an element like an icon, you should consult a screenshot
  to determine the coordinates of the element before moving the
  cursor.\n* If you tried clicking on a program or link but it failed
  to load, even after waiting, try adjusting your cursor position so
  that the tip of the cursor visually falls on the element that you
  want to click.\n* Make sure to click any buttons, links, icons, etc
  with the cursor tip in the center of the element. Don't click boxes
  on their edges.", "parameters": {"properties": {"action":
  {"description": "The action to perform. The available actions
  are:\n* `key`: Performs key down presses on the arguments passed in
  order, then performs key releases in reverse order.\n* `type`: Type
  a string of text on the keyboard.\n* `mouse_move`: Move the cursor
  to a specified (x, y) pixel coordinate on the screen.\n*
  `left_click`: Click the left mouse button at a specified (x, y)
  pixel coordinate on the screen.\n* `left_click_drag`: Click and drag
  the cursor to a specified (x, y) pixel coordinate on the screen.\n*
  `right_click`: Click the right mouse button at a specified (x, y)
  pixel coordinate on the screen.\n* `middle_click`: Click the middle
  mouse button at a specified (x, y) pixel coordinate on the
  screen.\n* `double_click`: Double-click the left mouse button at a
  specified (x, y) pixel coordinate on the screen.\n* `triple_click`:
  Triple-click the left mouse button at a specified (x, y) pixel
  coordinate on the screen (simulated as double-click since it's the
  closest action).\n* `scroll`: Performs a scroll of the mouse scroll
  wheel.\n* `hscroll`: Performs a horizontal scroll (mapped to regular
  scroll).\n* `wait`: Wait specified seconds for the change to
  happen.\n* `terminate`: Terminate the current task and report its
  completion status.\n* `answer`: Answer a question.\n* `back`: Go
  back to the previous page.\n* `home`: Go back to the task's starting
  page.", "enum": ["key", "type", "mouse_move", "left_click",
  "left_click_drag", "right_click", "middle_click", "double_click",
  "triple_click", "scroll", "hscroll", "wait", "terminate", "answer",
  "back", "home"], "type": "string"}, "keys": {"description":
  "Required only by `action=key`.", "type": "array"}, "text":
  {"description": "Required only by `action=type` and
  `action=answer`.", "type": "string"}, "coordinate": {"description":
  "(x, y): The x (pixels from the left edge) and y (pixels from the
  top edge) coordinates to move the mouse to.", "type": "array"},
  "pixels": {"description": "The amount of scrolling to perform.
  Positive values scroll up, negative values scroll down. Required
  only by `action=scroll` and `action=hscroll`.", "type": "number"},
  "time": {"description": "The seconds to wait. Required only by
  `action=wait`.", "type": "number"}, "status": {"description": "The
  status of the task. Required only by `action=terminate`.", "type":
  "string", "enum": ["success", "failure"]}}, "required": ["action"],
  "type": "object"}}}
</tools>

For each function call, return a json object with function name and
  arguments within <tool_call></tool_call> XML tags:
<tool_call>
{"name": <function-name>, "arguments": <args-json-object>}
</tool_call>

# Response format

Response format for every step:
1) Thought: one concise sentence explaining the next move (no
  multi-step reasoning).
2) Action: a short imperative describing what to do.
3) A single <tool_call>...</tool_call> block containing only the JSON:
  {"name": <function-name>, "arguments": <args-json-object>}.

Rules:
- Output exactly in the order: Thought, Action, <tool_call>.
- Be brief: one sentence for Thought, one for Action.
- Do not output anything else outside those three parts.
- If finishing, use computer_use with action=terminate in the tool
  call.

[user turn, after the current screenshot]
The user query: {instruction}
(For information questions, return the answer with the `answer` action
  (text='...') before terminating.)
Your step budget is limited: if the goal is already met,
  answer/terminate immediately; never repeat an action that did not
  change the page.
Task progress (You have done the following operation on the current
  page): {steps}
\end{verbatim}}
{\footnotesize
\begin{verbatim}
You are a GUI agent. You are given a task and your action history,
  with screenshots. You need to perform the next action to complete
  the task.

## Output Format
```
Thought: ...
Action: ...
```

## Action Space
click(start_box='<|box_start|>(x1,y1)<|box_end|>')
long_press(start_box='<|box_start|>(x1,y1)<|box_end|>', time='')
type(content='') # If you want to submit your input, use "\n" at the
  end of `content`.
scroll(start_box='<|box_start|>(x1,y1)<|box_end|>',
  end_box='<|box_start|>(x3,y3)<|box_end|>')
press_home()
press_back()
open_app(content='') # Open an app specified by `content`.
finished(content='') # Submit the task regardless of whether it
  succeeds or fails.
answer(content='') # Answer user's question.

## Note
- Use English in `Thought` and `Action` part.
- Write a small plan and finally summarize your next action (with its
  target element) in one sentence in `Thought` part.

## User Instruction
{instruction}


## Information tasks
For information-seeking tasks (questions asking how many / how long /
  what / which / to answer with ...), return the requested information
  using answer(content='...') -- do NOT terminate with an empty
  finished(). Only call finished() for non-question tasks.
\end{verbatim}}
{\footnotesize
\begin{verbatim}
# Tools

You may call one or more functions to assist with the user query.

You are provided with function signatures within <tools></tools> XML
  tags:
<tools>
{"type": "function", "function": {"name": "mobile_use", "description":
  "Use a touchscreen to interact with a mobile device, and take
  screenshots.\n* This is an interface to a mobile device with
  touchscreen. You can perform actions like clicking, typing, swiping,
  etc.\n* Some applications may take time to start or process actions,
  so you may need to wait and take successive screenshots to see the
  results of your actions.\n* The screen's resolution is 999x999.\n*
  Make sure to click any buttons, links, icons, etc with the cursor
  tip in the center of the element. Don't click boxes on their edges
  unless asked.", "parameters": {"properties": {"action":
  {"description": "The action to perform. The available actions
  are:\n* `click`: Click the point on the screen with coordinate (x,
  y).\n* `long_press`: Press the point on the screen with coordinate
  (x, y) for specified seconds.\n* `swipe`: Swipe from the starting
  point with coordinate (x, y) to the end point with coordinates2 (x2,
  y2).\n* `type`: Input the specified text into the activated input
  box.\n* `answer`: Output the answer.\n* `system_button`: Press the
  system button.\n* `open`: Open an app on the device.\n* `wait`: Wait
  specified seconds for the change to happen.\n* `terminate`:
  Terminate the current task and report its completion status.",
  "enum": ["click", "long_press", "swipe", "type", "answer",
  "system_button", "open", "wait", "terminate"], "type": "string"},
  "coordinate": {"description": "(x, y): The x (pixels from the left
  edge) and y (pixels from the top edge) coordinates to move the mouse
  to. Required only by `action=click`, `action=long_press`, and
  `action=swipe`.", "type": "array"}, "coordinate2": {"description":
  "(x, y): The x (pixels from the left edge) and y (pixels from the
  top edge) coordinates to move the mouse to. Required only by
  `action=swipe`.", "type": "array"}, "text": {"description":
  "Required only by `action=type`, `action=answer`, and
  `action=open`.", "type": "string"}, "time": {"description": "The
  seconds to wait. Required only by `action=long_press` and
  `action=wait`.", "type": "number"}, "button": {"description": "Back
  means returning to the previous interface, Home means returning to
  the desktop, Menu means opening the application background menu, and
  Enter means pressing the enter. Required only by
  `action=system_button`", "enum": ["Back", "Home", "Menu", "Enter"],
  "type": "string"}, "status": {"description": "The status of the
  task. Required only by `action=terminate`.", "type": "string",
  "enum": ["success", "failure"]}}, "required": ["action"], "type":
  "object"}}}
</tools>

For each function call, return a json object with function name and
  arguments within <tool_call></tool_call> XML tags:
<tool_call>
{"name": <function-name>, "arguments": <args-json-object>}
</tool_call>

# Response format

Response format for every step:
1) Thought: one concise sentence explaining the next move (no
  multi-step reasoning).
2) Action: a short imperative describing what to do.
3) A single <tool_call>...</tool_call> block containing only the JSON:
  {"name": <function-name>, "arguments": <args-json-object>}.

Rules:
- Output exactly in the order: Thought, Action, <tool_call>.
- Be brief: one sentence for Thought, one for Action.
- Do not output anything else outside those three parts.
- If finishing, use mobile_use with action=terminate in the tool call.

[user turn, after the current screenshot]
The user query: {instruction}
Task progress (You have done the following operation on the current
  device): {steps}
\end{verbatim}}

\paragraph{Judge.}
The judge receives the system prompt below, an evaluator prompt followed by a block of verification rules, and a user message that gives the task, every action of the episode in order with the screenshot after up to 48 uniformly sampled steps, and the final state, in the format shown after it.
{\footnotesize
\begin{verbatim}
You are an expert evaluator for mobile agent task completion. Your
  role is to determine whether a given task has been successfully
  completed based on the provided trajectory evidence.

## Evaluation Guidelines:

1. **Outcome-Focused**: Judge based on whether the primary objective
  was achieved, not the path taken.

2. **Use All Provided Evidence**: Base your judgment on all available
  information -- screenshots, actions, agent reasoning, and UI element
  lists. Do not assume information beyond what is provided.

3. **Mid-Trajectory Success**: The task may be completed in an
  intermediate step rather than the final one. Additional actions
  after completion do not invalidate success, unless they explicitly
  undo it.

4. **Corrective Actions**: If the agent made mistakes but corrected
  them and achieved the goal, that counts as success.

5. **UI State Indicators**: Pay attention to visual indicators like
  selected tabs, checkboxes, highlighted items, and confirmation
  messages.

Be balanced in your judgment - avoid being overly strict (missing true
  successes) or overly lenient (accepting failures).

VERIFICATION REQUIREMENTS (v1.1). Apply each rule only when the task
  involves
its situation; otherwise judge under the existing guidelines
  unchanged. When a
rule applies and its required evidence is absent or ambiguous, mark
  failure.

R1 -- State commitment. When the task requires creating, changing, or
submitting persistent state (saving an edit, placing an order,
  posting,
updating a status or setting), success requires evidence that the
  committing
action occurred (e.g., Save / Submit / Place Order / Post clicked)
  and, where
visible, the post-commit state (confirmation message, updated value
  shown
outside the editor, order-confirmation page). Entered-but-uncommitted
  input --
text typed into a field, items sitting in a cart, an edit visible only
  inside
an editor -- is not completion.

R2 -- Empty results. If the final answer or state amounts to "none /
  no
results / empty," verify from the evidence that the query or filter
  was
correctly constructed: correct field, correct spelling and wording of
  the
value, correct date range or option actually applied and visible. An
  empty
result produced by a malformed, misspelled, mis-scoped, or wrong-field
  query
is a failure even though the page truthfully shows no results. If
  query
correctness cannot be verified from the evidence, do not accept the
  empty
result as success.

R3 -- Selection constraints. When the task specifies which item among
candidates -- a superlative or ordering (latest, newest, first,
  cheapest,
most X) or a uniquely identifying property -- verify the selected item
satisfies that constraint with visible evidence (sort order applied,
  the
relevant dates/prices/attributes shown). Membership in the right set
  is not
sufficient. When the task prescribes a specific method or path (e.g.,
  browse
a stated category, use a stated page or tool), the prescribed path
  must have
been used; an alternative route is a failure even if it reached a
similar-looking result.

R4 -- Aggregate completeness. When the task asks for an aggregate or
  extremum
over a collection (a range, maximum, minimum, count, total), verify
  the
evidence covers the full collection -- pagination exhausted or a
  total/count
indicator visible -- not only the first page or a partial view.
  Answers
extrapolated from a visibly partial collection are failures.
\end{verbatim}}
{\footnotesize
\begin{verbatim}
## Task
{task}

## Agent Trajectory
The following shows the agent's execution trajectory in chronological
  order. Each step's screenshot is placed immediately after its
  description.

### Step {step_num}{step_label}
**Action**: {action}
[Screenshot attached]

### Step {step_num}{step_label}
**Action**: {action}

### Step {step_num} (Final State)
[Screenshot attached]
{optional_ui}

## Your Judgment
Based on the trajectory above, determine if the task was successfully
  completed.

Reply in the following format:
Result: <1 for success, 0 for failure>
Confidence: <high/medium/low>
Reason: <brief explanation in 1-2 sentences>
\end{verbatim}}

\paragraph{Proposer.}
The proposer receives the system prompt below and a user message with the original instruction and, for every step, the agent's thought, its action and the screenshot it saw, followed by the screenshot after the last action.
{\footnotesize
\begin{verbatim}
You are auditing a web-agent trajectory that FAILED its task. You will
  see the ORIGINAL task and, for each
step, the screenshot the agent saw BEFORE acting, its thought and its
  action (click(start_box='(x,y)') with x,y
in 0-1000 relative screen coordinates; type(content=...); scroll(...);
  press_back(); finished(...); answer(...)).

Your job is hindsight relabeling: decide whether some PREFIX of this
  trajectory correctly completed a coherent
task of its own -- a SUBTASK or PREREQUISITE of the original task
  (e.g. "open the issues page of repository X",
"search the store for 'wireless mouse' and open the results",
  "navigate to the settings page of forum F") --
even though the original task was not completed.

Rules for a valid proposal:
1. The pseudo task must be a subtask, prerequisite or strictly
  narrower version of the ORIGINAL task, on the same
   site and about the same objects; never an unrelated task, and never
     the original task itself.
2. It must be COMPLETED by the agent's actions up to some step k
  (0-based): the screenshot the agent saw at step
   k+1 (i.e. after acting at step k) must show the task's result.
     Choose the EARLIEST such k; do not include
   later steps that wander, loop or undo progress.
3. It must be non-trivial: at least one navigation or typing action
  beyond loading the start page; "scroll the
   page", "open the homepage" or "click around" are not tasks.
4. It must be stated like a user instruction (one sentence,
  imperative, concrete names/values), and be
   verifiable from the final screenshot alone.
5. If no prefix satisfies all of the above, answer valid=false.
6. The pseudo task must END ON A RESULTING PAGE OR STATE produced by a
  page-changing action (a search
   results page, a product / repository / report page, a submitted
     form, a changed setting). Text typed
   into a field that was not submitted, an opened menu, or a
     highlighted element is NOT a completed task.
7. The pseudo task must describe a SUCCESSFUL outcome that a user
  would ask for. An error message, a
   validation failure, an empty or 'no results' page, or an attempt
     that did not go through is NOT a
   completed task.

Respond with ONLY a JSON object:
{"valid": true|false, "pseudo_task": "<instruction or empty>",
  "end_step": <k or -1>,
 "relation": "subtask"|"prerequisite"|"narrower"|"none", "reason":
   "<one sentence>"}
\end{verbatim}}
{\footnotesize
\begin{verbatim}
ORIGINAL TASK (failed): {instruction}

The trajectory has {n} steps.
--- Step 0 (screenshot before acting below)
Thought: {thought 0}
{action 0}
[screenshot 0]
--- Step 1 (screenshot before acting below)
...
--- Screenshot after the last action (step {n-1}):
[screenshot n]
\end{verbatim}}

\paragraph{Verifier.}
The verifier receives only the proposed instruction, the actions of the prefix and the screenshot after its last step.
{\footnotesize
\begin{verbatim}
You are a strict verifier for a web-agent task. You will see a TASK,
  the list of actions the agent took, and
the FINAL screenshot after those actions. Decide whether the final
  screenshot shows that the task has been
completed as stated (correct page, correct object, requested state
  reached). Be conservative: if the
screenshot does not by itself demonstrate completion, or the task is
  vague or trivial, answer false.

Respond with ONLY a JSON object: {"verified": true|false,
  "confidence": <0-1>, "reason": "<one sentence>"}

[user turn]
TASK: {pseudo task}

ACTIONS TAKEN:
0: Action: {action 0}
1: Action: {action 1}
...

FINAL SCREENSHOT:
[screenshot after step k]
\end{verbatim}}

\paragraph{Memory baselines.}
AWM-online induces its workflows with the authors' instruction, reproduced first, and a one-shot example rewritten in our agents' action vocabulary, reproduced in its \uitars form.
The induced workflow file enters the agent's prompt as the block shown last, with the action-space note appended to the file.
{\footnotesize
\begin{verbatim}
Given a list of web navigation tasks, your task is to extract the
  common workflows to solve these tasks.
Each given task contains a natural language instruction, and a series
  of actions to solve the task. You need to find the repetitive subset
  of actions across multiple tasks, and extract each of them out as a
  workflow.
Each workflow should be a commonly-reused sub-routine of the tasks. Do
  not generate similar or overlapping workflows. Each workflow should
  have at least two steps. Represent the non-fixed elements (input
  text, button strings) with descriptive variable names as shown in
  the example.
Keep the values of invariant elements, e.g., id of "Search" or
  "Customers", as they will share and stay invariant across tasks.
Try to generate as many workflows that can cover all the tasks in the
  input list.
\end{verbatim}}
{\footnotesize
\begin{verbatim}
## Concrete Examples

Query: What is the date of my first purchase on this store?
Actions:
<think>
To find the first purchase I need the order history, which lives under
  the account menu.
</think>
<action>
Click on the account menu in the top-right corner of the page.
</action>

<think>
The order history is the "My Orders" entry in the account page's left
  sidebar.
</think>
<action>
Click the "My Orders" entry in the left sidebar.
</action>

## Summary Workflows

Workflow 1: Open the account's order history
<think>
To reach anything about past orders I first open the account menu in
  the top-right corner.
</think>
<action>
Click on the account menu in the top-right corner of the page.
</action>

<think>
From the account page the order history is the "My Orders" entry in
  the left sidebar.
</think>
<action>
Click the "My Orders" entry in the left sidebar.
</action>

Workflow 2: Search the store for a product
<think>
I put the product I am looking for, {product-name}, into the site's
  search box.
</think>
<action>
Type {product-name} into the search box and press Enter.
</action>
\end{verbatim}}
{\footnotesize
\begin{verbatim}
## Workflows
These sub-routines were induced from earlier tasks on this website
  that were completed successfully. Reuse the matching one,
  substituting this task's own values for the {{placeholders}} -- read
  those from the current instruction and the current screen, never
  from a workflow.
{workflows}

[action-space note appended to every induced workflow file, UI-TARS
  form]

(Write each step as the action you would take; keep the page's own
  button and field names, and substitute this task's values for every
  {placeholder}.)

[Qwen3-VL form]

(Each step is a computer_use tool call, exactly as you emit them;
  substitute this task's values for every {placeholder}.)
\end{verbatim}}
Darwinian Memory shows the agent its retrieved units in the block below.
{\footnotesize
\begin{verbatim}
## Retrieved memory
Each unit below was recorded when an earlier task with a similar goal
  succeeded from a similar starting state. Replay the matching steps
  where they still fit what you see; this task's own values must come
  from the current instruction and screen.
{units}

[one unit]
### Goal: {goal}
Precondition: {precond}
Actions:
{steps}
\end{verbatim}}

\end{document}